\documentclass[11pt,a4paper]{article}
\usepackage[hyperref]{acl2021}
\usepackage{times}
\usepackage{latexsym}

\usepackage{microtype}

\usepackage{kotex}

\usepackage[para]{threeparttable}
\usepackage{booktabs}
\usepackage{tabularx}
\usepackage{makecell}
\usepackage{amsmath} 
\usepackage{amssymb} 
\usepackage[toc,page]{appendix}
\usepackage{float}
\usepackage{amsthm} 
\usepackage{graphicx} 
\usepackage{array} 
\usepackage{multirow}
\usepackage{csquotes} 
\usepackage{algorithmic}
\usepackage{hyperref}
\usepackage{multicol}
\usepackage{balance}

\usepackage[para]{threeparttable}
\usepackage{booktabs}
\usepackage{tabularx}
\usepackage[ruled,vlined]{algorithm2e}

\aclfinalcopy 

\title{NeuralWOZ: Learning to Collect Task-Oriented Dialogue \\ via Model-Based Simulation}

\author{
    Sungdong Kim$^{1,2}$ \quad Minsuk Chang$^{1,2}$ \quad Sang-Woo Lee$^{1,2}$ \\
    NAVER AI Lab$^{1}$  NAVER Clova$^{2}$\\
    \texttt{\{sungdong.kim, minsuk.chang, sang.woo.lee\}@navercorp.com}
}

\date{}

\begin{document}
\maketitle

\begin{abstract}
We propose NeuralWOZ, a novel dialogue collection framework that uses model-based dialogue simulation. NeuralWOZ has two pipelined models, Collector and Labeler. Collector generates dialogues from (1) user's goal instructions, which are the user context and task constraints in natural language, and (2) system's API call results, which is a list of possible query responses for user requests from the given knowledge base. Labeler annotates the generated dialogue by formulating the annotation as a multiple-choice problem, in which the candidate labels are extracted from goal instructions and API call results. We demonstrate the effectiveness of the proposed method in the zero-shot domain transfer learning for dialogue state tracking. In the evaluation, the synthetic dialogue corpus generated from NeuralWOZ achieves a new state-of-the-art with improvements of 4.4\% point joint goal accuracy on average across domains, and improvements of 5.7\% point of zero-shot coverage against the MultiWOZ 2.1 dataset.\footnote{The code is available at \href{https://github.com/naver-ai/neuralwoz}{github.com/naver-ai/neuralwoz}.}
\end{abstract}

\section{Introduction}
\begin{figure}[t] 
\centering
\includegraphics[width=0.47\textwidth]{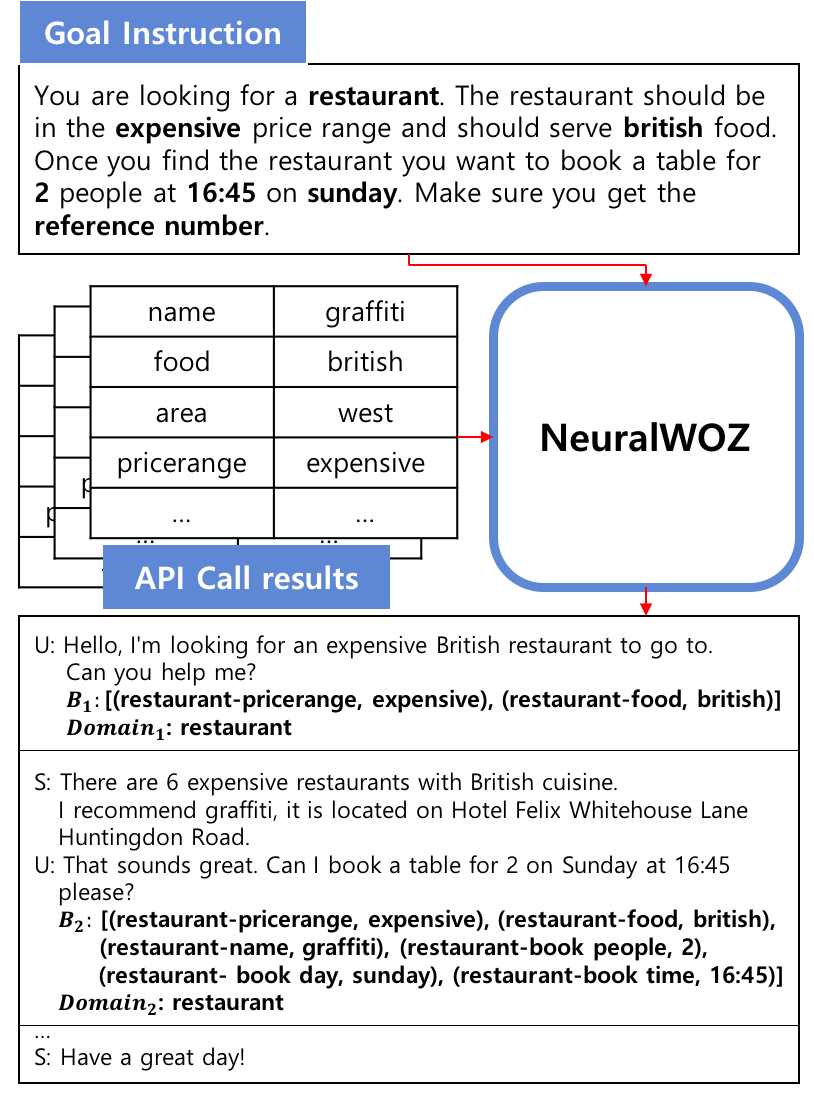}
\caption{Overview of NeuralWOZ. The NeuralWOZ takes goal instruction for the user side (U) and API call results for the system side (S) to synthesize dialogue. First, it generates dialogue from the inputs and then labels dialogue state ($B_t$) and active domain ($Domain_t$) by turn $t$ on the dialogue.}
\label{fig:neuralwoz_overview}
\end{figure}

For a task-oriented dialogue system to be scalable, the dialogue system needs to be able to quickly adapt and expand to new scenarios and domains. However, the cost and effort in collecting and annotating an expanding dataset is not only labor-intensive but also proportional to the size and variety of the unseen scenarios.

There are three types of dialogue system expansions. (1) The simplest expansion is the addition of new instances in the knowledge base (KB) under the identical schema. For example, the addition of newly opened restaurants in the KB of restaurant domain falls under this category. (2) A slightly more complicated expansion involves modifications to the KB schema, and possibly the related instances. For example, additions of new constraint types to access the KB due to the change in needs of the user often require a restructuring of the KB. If a dialogue system built with only restaurant search in mind observes user's requests about not only ``restaurant location" and but also ``traffic information" for navigating, the system now needs a new knowledge base including the additional different domain. (3) The most complex expansion is the one that expands across multiple domains. For example, imagine an already built dialogue system supported restaurant and hotel reservation domains, but now needs to expand to points of interest or other domains. It is difficult to expand to new domain without collecting new data instances and building a new knowledge base, if the schema between the source (restaurant and hotel in this case) and target domain (point of interest) look different.

To support development of scalable dialogue systems, we propose NeuralWOZ, a model-based dialogue collection framework. NeuralWOZ uses goal instructions and KB instances for synthetic dialogue generation. NeuralWOZ mimics the mechanism of a Wizard-of-Oz \cite{kelley1984iterative, dahlback1993wizard} and Figure \ref{fig:neuralwoz_overview} illustrates our approach. NeuralWOZ has two neural components, Collector and Labeler. Collector generates a dialogue by using the given goal instruction and candidate relevant API call results from the KB as an input. Labeler annotates the generated dialogue with appropriate labels by using the schema structure of the dialogue domain as meta information. More specifically, Labeler selects the labels from candidate labels which can be obtained from the goal instruction and the API call results. As a result, NeuralWOZ is able to generate a dialogue corpus without training data of the target domain.

We evaluate our method for zero-shot domain transfer task~\cite{wu-etal-2019-transferable, campagna-etal-2020-zero} to demonstrate the ability to generate corpus for unseen domains, when no prior training data exists. In dialogue state tracking (DST) task with MultiWOZ 2.1~\cite{eric2019multiwoz}, the synthetic data generated with NeuralWOZ achieves 4.4\% point higher joint goal accuracy and 5.7\% point higher zero-shot coverage than the existing baseline. Additionally, we examine few-shot and full data augmentation tasks using both training data and synthetic data. We also illustrate how to collect synthetic data beyond MultiWOZ domains, and discuss the effectiveness of the proposed approach as a data collection strategy.

Our contributions are as follows:
\begin{itemize}
    \setlength\itemsep{0em}
    \item NeuralWOZ, a novel method for generating dialogue corpus using goal instruction and knowledge base information
    \item New state-of-the-art performance on the zero-shot domain transfer task 
    \item Analysis results highlighting the potential synergy of using the data generated from NeuralWOZ together with human-annotated data 
\end{itemize}

\section{Related Works}
\subsection{Wizard-of-Oz}
Wizard-of-Oz (WOZ) is a widely used approach for constructing dialogue data~\cite{henderson-etal-2014-second, henderson2014third, el-asri-etal-2017-frames, eric2017keyvalue, budzianowski-etal-2018-multiwoz}. It works by facilitating a role play between two people. 
``User'' utilizes a goal instruction that describes the context of the task and details of request and ``system'' has access to a knowledge base, and query results from the knowledge base. They take turns to converse, while the user makes requests one by one following the instructions, the system responds according to the knowledge base, and labels user's utterances.

\subsection{Synthetic Dialogue Generation}
Other studies on dialogue datasets use the user simulator-based data collection approaches \cite{schatzmann-etal-2007-agenda, li2017user, bordes2017learning, shah2018building, zhao2018zero, shah2018building, campagna-etal-2020-zero}. They define domain schema, rules, and dialogue templates to simulate user behavior under certain goals. The ingredients to the simulation are designed by developers and the dialogues are realized by predefined mapping rules or paraphrasing by crowdworkers.

If a training corpus for the target domain exists, neural models that synthetically generates dialogues can augment the training corpus~\cite{hou-etal-2018-sequence,yoo2019data}. For example, \citet{yoo-etal-2020-variational} introduce Variational Hierarchical Dialog Autoencoder (VHDA), where hierarchical latent variables exist for speaker identity, user's request, dialog state, and utterance. They show the effectiveness of their model on single-domain DST tasks. SimulatedChat~\cite{mohapatra2020simulated} also uses goal instruction for dialogue augmentation. Although it does not solve zero-shot learning task with domain expansion in mind, we run auxiliary experiments to compare with NeuralWOZ, and the results are in the Appendix \ref{appendix.simulatedchat}.

\subsection{Zero-shot Domain Transfer}
In zero-shot domain transfer tasks, there is no data for target domain, but there exists plenty of data for other domains similar to target domain. Solving the problem of domain expansion of dialogue systems can be quite naturally reducted to solving zero-shot domain transfer. \citet{wu-etal-2019-transferable} conduct a landmark study on the zero-shot DST. They suggest a model, Transferable Dialogue State Generator (TRADE), which is robust to a new domain where few or no training data for the domain exists. \citet{kumar2020madst} and \citet{li-etal-2021-zero} follow the same experimental setup, and we also compare NeuralWOZ in the same experiment setup. Abstract Transaction Dialogue Model (ATDM)~\cite{campagna-etal-2020-zero}, another method for synthesizing dialogue data, is another baseline for zero-shot domain transfer tasks we adopt. 
They use rules, abstract state transition, and templates to synthesize the dialogue, which is then fed into a model-based zero-shot learner. They achieved state-of-the-art in the task using the synthetic data on SUMBT \cite{lee-etal-2019-sumbt}, a pretrained BERT \cite{devlin-etal-2019-bert} based DST model.

\section{NeuralWOZ}
\begin{figure}[t] 
\centering
\includegraphics[width=0.47\textwidth]{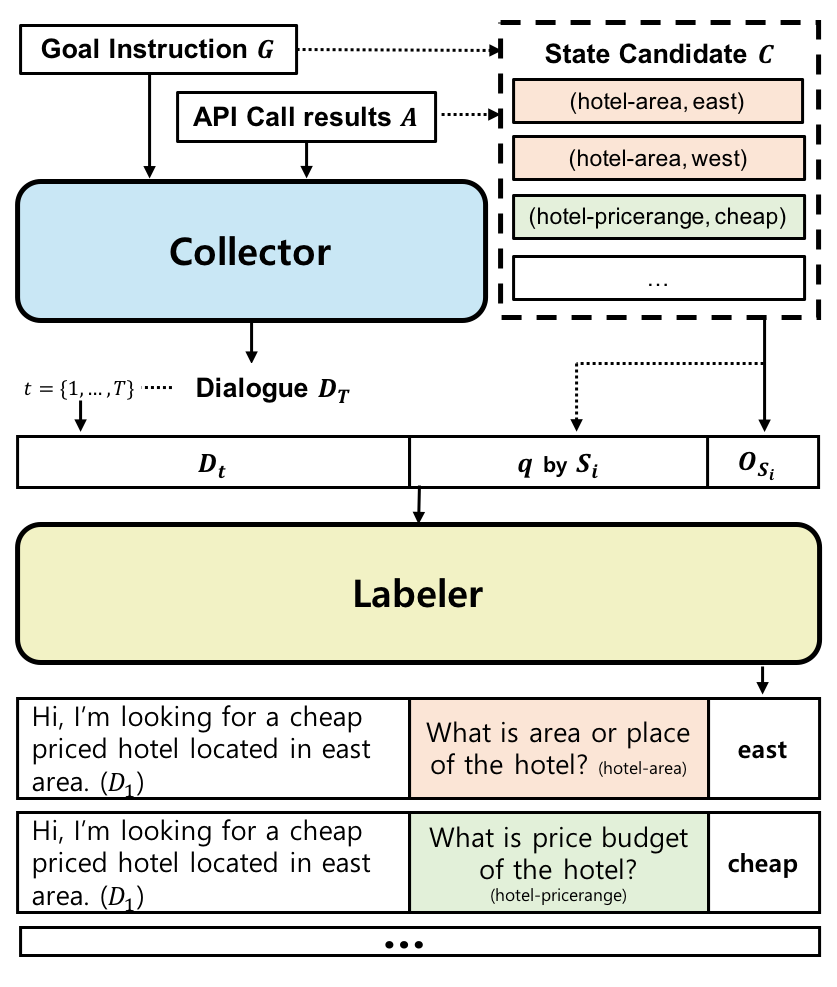}
\caption{Illustration of Collector and Labeler. Collector takes goal instruction $G$ and API call results $A$ as the input, and outputs dialogue $D_{T}$ which consists of $T$ turns. The state candidate $C$ is prepopulated from the $G$ and $A$ as a full set for labeling. Finally, Labeler takes its value's subset $O_{S_i}$ and question $q$ for each slot type $S_i$ and dialogue context $D_t$ from Collector, and chooses answer $\tilde{o}$ from the $O_{S_i}$.}
\label{fig:collector_labeler}
\end{figure}

In this section, we describe the components of NeuralWOZ in detail, and how they interact with each other. 
Figure \ref{fig:collector_labeler} illustrates the input and output of two modules in NeuralWOZ.
The synthetic corpus, which Collector and Labeler made, are used for the training of the DST baselines, TRADE~\cite{wu-etal-2019-transferable} and SUMBT~\cite{lee-etal-2019-sumbt} in our experiments.

\subsection{Problem Statement}
\medskip \noindent
\textbf{Domain Schema} In task-oriented dialogues, there are two slot types; $informable$ and $requestable$ slots ~\cite{henderson-etal-2014-second, budzianowski-etal-2018-multiwoz}. The $informable$ slots are the task constraints to find relevant information from user requests, for example,  ``restaurant-pricerange'', ``restaurant-food'', ``restaurant-name'', and ``restaurant-book people'' in Figure~\ref{fig:neuralwoz_overview}. The $requestable$ slots are the additional details of user requests, like ``reference number'' and ``address'' in Figure \ref{fig:neuralwoz_overview}. Each slot $S$ can have its corresponding value $V$ in a scenario. In multi-domain scenarios, each domain has a knowledge base $KB$, which consists of slot-value pairs corresponding to its domain schema. The API call results in Figure~\ref{fig:neuralwoz_overview} are the examples of the $KB$ instances of the restaurant domain.

\medskip \noindent
\textbf{Goal Instruction} The goal instruction, $G$, is a natural language text describing constraints of user behavior in the dialogue $D$ including informable and requestable slots. The paragraph consists of four sentences at the top of Figure \ref{fig:neuralwoz_overview} is an example. We define a set of informable slot-value pairs that explicitly expressed on the $G$ as $C^G$, which we formally define as $C^G = \{(S_i^G, V_i^G) \mid 1 \le i \le |C^G|, S_i^G \in informable\}$.
(``restaurant-pricerange'', ``expensive'') and (``restaurant-food'', ``british'') are examples of the elements of $C^G$ (Figure~\ref{fig:neuralwoz_overview}).

\medskip \noindent
\textbf{API Call Results} The API call results, $A$, are corresponding query results of the $C^G$ from KB. We formally define $A = \{a_i \mid 1 \le i \le |A|, a_i \in KB\}$. Each $a_i$ is associated with its domain, $domain_{a_i}$, and with slot-value pairs,  $C^{a_i} = \{(S_k^{a_i}, V_k^{a_i}) \mid 1 \le k \le |C^{a_i}|\}$. A slot $S_k^{a_i}$ can be either informable or requestable slot. For example, the restaurant instance, ``graffiti'' in Figure~\ref{fig:neuralwoz_overview}, is a query result from (``restaurant-pricerange'', ``expensive'') and (``restaurant-food'', ``british'') described in the goal instruction.

\medskip \noindent
\textbf{State Candidate} We define informable slot-value pairs that are not explicit in $G$ but accessible by $A$ in $D$ as $C^A = \{(S_i^A, V_i^A) \mid 1 \le i \le |C^A|, S_i^A \in informable\}$. It contains all informable slot-value pairs from $C^{a_1}$ to $C^{a_{|A|}}$. The elements of $C^A$ are likely to be uttered by summaries of current states or recommendations of KB instances by the system side in $D$. The system utterance of the second turn in Figure~\ref{fig:neuralwoz_overview} is an example (``I recommend graffiti.''). In this case, the slot-value pair (``restaurant-name'', ``graffiti'') can be obtained from the $A$, not from the $G$.
Finally, state candidate $C$ is the union of $C^G$ and $C^A$. It is a full set of the dialogue state for the dialogue $D$ from given $G$ and $A$. Thus, it can be used as label candidates of dialogue state tracking annotation.

\subsection{Collector}
Collector is a sequence-to-sequence model, which takes a goal instruction $G$ and API call results $A$ as the input and generates dialogue $D_T$. The generated dialogue $D_T = (r_1, u_1, ..., r_T, u_T)$ is the sequence of system response $r$ and user utterance $u$. They are represented by $N$ tokens $(w_1, ..., w_N)$\footnote{Following \citet{hosseiniasl2020simple}, we also utilize role-specific special tokens \texttt{<system>} and \texttt{<user>} for the $r$ and $u$ respectively.}.
\begin{equation*}
    p(D_T|G, A) = \prod_{i=1}^N p(w_i|w_{<i}, G, A)
\end{equation*}
We denote the input of Collector as $\texttt{<s>} \oplus G \oplus \texttt{</s>} \oplus A$, where the $\oplus$ is concatenate operation. The \texttt{<s>} and \texttt{</s>} are special tokens to indicate start and seperator respectively. The tokenized natural language description of $G$ is directly used as the tokens. The $A$ takes concatenation of each $a_i$ ($a_1 \oplus \cdots \oplus a_{|A|}$)\footnote{we limit the $|A|$ to a maximum 3}. For each $a_i$, we flatten the result to the token sequence, $\texttt{<domain>} \oplus domain_{a_i} \oplus \texttt{<slot>} \oplus S_1^{a_i} \oplus V_1^{a_i} \oplus \cdots \oplus \texttt{<slot>} \oplus S_{|C^{a_i}|}^{a_i} \oplus V_{|C^{a_i}|}^{a_i}$. The \texttt{<domain>} and \texttt{<slot>} are other special tokens as separators.
The objective function of Collector is
\begin{equation*}
    \mathcal{L}_C = -\frac{1}{M_C} \sum_{j=1}^{M_C} \sum_{i=1}^{N_j} \log p(w_i^j|w_{<i}^j, G^j, A^j).
\end{equation*}
Our Collector model uses the transformer architecture~\cite{vaswani2017attention} initialized with pretrained BART \cite{lewis-etal-2020-bart}. Collector is trained using negative log-likelihood loss, where $M_C$ is the number of training dataset for Collector and $N_j$ is target length of the $j$-th instance. Following \citet{lewis-etal-2020-bart}, label smoothing is used during the training with the smoothing parameter of 0.1.

\subsection{Labeler}
We formulate labeling as a multiple-choice problem. Specifically, Labeler takes a dialogue context $D_t = (r_1, u_1, ..., r_t, u_t)$, question $q$, and a set of answer options $O = \{o_1, o_2, ..., o_{|O|}\}$, and selects one answer $\tilde{o} \in O$. Labeler encodes the inputs for each $o_i$ separately, and $s_{o_i} \in \mathbb{R}^1$ is the corresponding logit score from the encoding. Finally, the logit score is normalized via softmax function over the answer option set $O$.
\begin{equation*}
\begin{aligned}
    &p(o_i|D_t, q, O) = \frac{\exp(s_{o_i})}{\sum_j^{|O|}\exp(s_{o_j})}, \\
    &s_{o_i} = Labeler(D_t, q, o_i), \forall i.
\end{aligned}
\end{equation*}
The input of Labeler is a concatenation of $D_t$, $q$, and $o_i$, $\texttt{<s>} \oplus D_t \oplus \texttt{</s>} \oplus q \oplus \texttt{</s>} \oplus o_i \oplus \texttt{</s>}$, with special tokens.
For labeling dialogue states to $D_t$, we use the slot description for each corresponding slot type, $S_i$, as the question, for example, ``what is area or place of hotel?'' for ``hotel-area'' in Figure~\ref{fig:collector_labeler}. We populate corresponding answer options $O_{S_i} = \{V_j | (S_j, V_j) \in C, S_j = S_i \}$ from the state candidate set $C$. There are two special values, $Dontcare$ to indicate the user has no preference and $None$ to indicate the user is yet to specify a value for this slot~\cite{henderson-etal-2014-second, budzianowski-etal-2018-multiwoz}. We include these values in the $O_{S_i}$.

For labeling the active domain of $D_t$, which is the domain at $t$-th turn of $D_t$, we define domain question, for example ``what is the domain or topic of current turn?", for $q$ and use predefined domain set {$O_{domain}$ as answer options. In MultiWOZ, $O_{domain}$ = \{``Attraction'', ``Hotel'', ``Restaurant'', ``Taxi'', ``Train''\}.}

Our Labeler model employs a pretrained RoBERTa model~\cite{liu2019roberta} as the initial weight. Dialogue state and domain labeling are trained jointly based on the multiple choice setting. Preliminary result shows that the imbalanced class problem is significant in the dialogue state labels. Most of the ground-truth answers is $None$ given question\footnote{The number of $None$ in the training data is about 10 times more than the number of others}. Therefore, we revise the negative log-likelihood objective to weight other (not-$None$) answers by multiplying a constant $\beta$ to the log-likelihood when the answer of training instance is not $None$. The objective function of Labeler is 
\begin{equation*}
\begin{aligned}
    &\mathcal{L}_L = -\frac{1}{M_L} \sum_{j=1}^{M_L}\sum_{t=1}^T\sum_{i=1}^{N_q} \mathcal{L}_{t, i}^j \\
    &\mathcal{L}_{t, i}^j= 
    \begin{cases}
        \beta \log p(\tilde{o}_{t, i}^j|D_t^j, q_i^j, O_i^j),& \text{if } \tilde{o}_{t, i}^j \ne \text{$None$}\\
        \log p(\tilde{o}_{t, i}^j|D_t^j, q_i^j, O_i^j),              & \text{otherwise}
    \end{cases}
\end{aligned}
\end{equation*}
, where $\tilde{o}_{t, i}^j$ denotes the answer of $i$-th question for $j$-th training dialogue at turn $t$, the $N_q$ is the number of questions, and $M_L$ is the number of training dialogues for Labeler.
We empirically set $\beta$ to a constant $5$.

\subsection{Synthesizing a Dialogue}
We first define goal template $\mathcal{G}$.\footnote{In \citet{budzianowski-etal-2018-multiwoz}, they also use templates like ours when allocating goal instructions to the user in the Wizard-of-Oz setup.} $\mathcal{G}$ is a delexicalized version of $G$ by changing each value $V_i^G$ expressed on the instruction to its slot $S_i^G$. For example, the ``expensive'' and ``british'' of goal instruction in Figure~\ref{fig:neuralwoz_overview} are replaced with ``restaurant-pricerange'' and ``restaurant-food'', respectively. As a result, domain transitions in $\mathcal{G}$ becomes convenient.

First, $\mathcal{G}$ is sampled from a pre-defined set of goal template. API call results $\mathcal{A}$, which correspond to domain transitions in $\mathcal{G}$, are randomly selected from the $KB$. Especially, we constrain the sampling space of $\mathcal{A}$ when the consecutive scenario among domains in $\mathcal{G}$ have shared slot values. For example, the sampled API call results for restaurant and hotel domain should share the value of ``area'' to support the following instruction ``I am looking for a hotel nearby the restaurant''.  $\mathcal{G}$ and $ \mathcal{A}$ are aligned to become $G_\mathcal{A}$. In other words, each value for $S_i^G$ in $\mathcal{G}$ is assigned using the corresponding values in $\mathcal{A}$.\footnote{Booking-related slots, e.g., the number of people, time, day, and etc., are randomly sampled for their values since they are independent of the $A$.} Then, Collector generates dialogue $\mathcal{D}$, of which the total turn number is $T$, given $G_\mathcal{A}$ and $\mathcal{A}$. More details are in Appendix \ref{appendix.synthesize}. Nucleus sampling \cite{holtzman2020curious} is used for the generation. 

We denote dialogue state and active domain at turn $t$ as $B_t$ and $domain_t$ respectively. The $B_t$, $\{(S_j, V_{j, t}) \mid 1 \le j \le J\}$, has $J$ number of predefined slots and their values at turn $t$. It means Labeler is asked $J$ (from slot descriptions) + 1 (from domain question) questions regarding dialogue context $\mathcal{D}_t$ from Collector. Finally, the output of Labeler is a set of dialogue context, dialogue state, and active domain at turn $t$ triples $\{(\mathcal{D}_1, B_1, domain_1), ..., (\mathcal{D}_T, B_T, domain_T)\}$.

\section{Experimental Setups}
\begin{table*}[t!]
    \centering
    \footnotesize
    \begin{threeparttable}
    \begin{tabular*}{0.988\textwidth}{c|c|ccccc|c}
        \toprule
Model & Training & Hotel & Restaurant & Attraction & Train & Taxi & Average \\
        \midrule
        \multirow{7}{*}{\textbf{TRADE}} & Full dataset & 50.5 / 91.4 & 61.8 / 92.7 & 67.3 / 87.6 & 74.0 / 94.0 & 72.7 / 88.9 & 65.3 / 89.8\\
        \cmidrule{2-8}
        & Zero-shot (\textit{Wu}) & 13.7 / 65.6 & 13.4 / 54.5 & 20.5 / 55.5 & 21.0 / 48.9 & 60.2 / 73.5 & 25.8 / 59.6  \\
        & Zero-shot (\textit{Campagna}) & 19.5 / 62.6 & 16.4 / 51.5 & 22.8 / 50.0 & 22.9 / 48.0 & 59.2 / 72.0 & 28.2 / 56.8  \\
        & Zero-shot + ATDM & \textbf{28.3} / 74.5 & 35.9 / 75.6 & 34.9 / 62.2 & 37.4 / 74.5 & 65.0 / 79.9 & 40.3 / 73.3 \\
        & Zero-shot + NeuralWOZ & 26.5 / \textbf{75.1} & \textbf{42.0} / \textbf{84.2} & \textbf{39.8} / \textbf{65.7} & \textbf{48.1} / \textbf{83.9} & \textbf{65.4} / \textbf{79.9} & \textbf{44.4} / \textbf{77.8} \\
        \cmidrule{2-8}
        & Zero-shot Coverage & 52.5 / 82.2 & 68.0 / 90.8 & 59.1 / 75.0 & 65.0 / 89.3 & 90.0 / 89.9 & 66.9 / 85.4 \\
        \midrule \midrule
        \multirow{6}{*}{\textbf{SUMBT}} & Full dataset & 51.8 / 92.2 & 64.2 / 93.1 & 71.1 / 89.1 & 77.0 / 95.0 & 68.2 / 86.0 & 66.5 / 91.1 \\
        \cmidrule{2-8}
        & Zero-shot & 19.8 / 63.3 & 16.5 / 52.1 & 22.6 / 51.5 & 22.5 / 49.2 & 59.5 / 74.9 & 28.2 / 58.2  \\
        & Zero-shot + ATDM & \textbf{36.3} / \textbf{83.7} & 45.3 / 82.8 & 52.8 / 78.9 & 46.7 / 84.2 & 62.6 / 79.4 & 48.7 / 81.8 \\
        & Zero-shot + NeuralWOZ & 31.3 / 81.7 & \textbf{48.9} / \textbf{88.4} & \textbf{53.0} / \textbf{79.0} & \textbf{66.9} / \textbf{92.4} & \textbf{66.7} / \textbf{83.9} & \textbf{53.4} / \textbf{85.1} \\
        \cmidrule{2-8}
        & Zero-shot Coverage & 60.4 / 88.6 & 76.2 / 95.0 & 74.5 / 88.7 & 86.9 / 97.3 & 97.8 / 97.6 & 79.2 / 93.4 \\
        \bottomrule
    \end{tabular*}
    \end{threeparttable}
    \caption{Experimental results of zero-shot domain transfer on the test set of MultiWOZ 2.1. Joint goal accuracy / slot accuracy are reported. The \textit{Wu} indicates original zero-shot scheme of the TRADE suggested by \citet{wu-etal-2019-transferable} and reproduced by \citet{campagna-etal-2020-zero}. The \textit{Campagna} indicates a revised version of the original by \citet{campagna-etal-2020-zero}. The + indicates the synthesized dialogue is used together for the training.}
    \label{table:zero-shot}
\end{table*}


\subsection{Dataset}
We use MultiWOZ 2.1 \cite{eric2019multiwoz} dataset\footnote{https://github.com/budzianowski/multiwoz} for our experiments. It is one of the largest publicly available multi-domain dialogue data and it contains 7 domains related to travel (attraction, hotel, restaurant, taxi, train, police, hospital), including about 10,000 dialogues. The MultiWOZ data is created using WOZ so it includes goal instruction per each dialogue and domain-related knowledge base as well. We train our NeuralWOZ using the goal instructions and the knowledge bases first. Then we evaluate our method on dialogue state tracking with and without synthesized data from the NeuralWOZ using five domains (attraction, restaurant, hotel, taxi, train) in our baseline, and follow the same preprocessing steps of ~\citet{wu-etal-2019-transferable, campagna-etal-2020-zero}. 

\subsection{Training NeuralWOZ}
We use the pretrained BART-Large~\cite{lewis-etal-2020-bart} for Collector and RoBERTa-Base~\cite{liu2019roberta} for Labeler. They share the same byte-level BPE vocab \cite{sennrich-etal-2016-neural} introduced by \citet{radford2019language}. We train the pipelined models using Adam optimizer~\cite{kingma2017adam} with learning rate 1e-5, warming up steps 1,000, and batch size 32. The number of training epoch is set to 30 and 10 for Collector and Labeler respectively.

For the training phase of Labeler, we use a state candidate set from ground truth dialogue states $B_{1:T}$ for each dialogue, not like the synthesizing phase where the options are obtained from goal instruction and API call results. We also evaluate the performance of Labeler itself like the training phase with validation data (Table \ref{table:responsibility}). Before training Labeler on the MultiWOZ 2.1 dataset, we pretrain Labeler on DREAM\footnote{The DREAM is a multiple-choice question answering dataset in dialogue and includes about 84\% of non-extractive answers.}~\cite{sun2019dream} to boost Labeler's performance. This is similar to coarse-tuning in~\citet{jin2019mmm}. The same hyper parameter setting is used for the pretraining.

For the zero-shot domain transfer task, we exclude dialogues which contains target domain from the training data for both Collector and Labeler. This means we train our pipelines for every target domain separately. We use the same seed data for training as ~\citet{campagna-etal-2020-zero} did in the few-shot setting. All our implementations are conducted on NAVER Smart Machine Learning (NSML) platform~\cite{sung2017nsml,kim2018nsml} using huggingface's transformers library~\cite{wolf-etal-2020-transformers}. The best performing models, Collector and Labeler, are selected by evaluation results from the validation set.

\subsection{Synthetic Data Generation}
We synthesize 5,000 dialogues for every target domain for both zero-shot and few-shot experiments\footnote{In \citet{campagna-etal-2020-zero}, the average number of synthesized dialogue over domains is 10,140.}, and 1,000 dialogues for full data augmentation. For zero-shot experiment, since the training data are unavailable for a target domain, we only use goal templates that contain the target domain scenario in the validation set similar to \citet{campagna-etal-2020-zero}. We use nucleus sampling in Collector with parameters top\_p ratio in the range $\{0.92, 0.98\}$ and temperature in the range $\{0.7, 0.9, 1.0\}$. It takes about two hours to synthesize 5,000 dialogues using one V100 GPU. More statistics is in Appendix \ref{appendix.stats}.

\subsection{Baselines}
We compare NeuralWOZ with baseline methods both zero-shot learning and data augmentation using MultiWOZ 2.1 in our experiments.
We use a baseline zero-shot learning scheme which does not use synthetic data~\cite{wu-etal-2019-transferable}.
For data augmentation, we use ATDM and VHDA.

ATDM refers to a rule-based synthetic data augmentation method for zero-shot learning suggested by \citet{campagna-etal-2020-zero}. It defines rules including state transitions and templates for simulating dialogues and creates about 10,000 synthetic dialogues per five domains in the MultiWOZ dataset. \citet{campagna-etal-2020-zero} feed the synthetic dialogues into zero-shot learner models to perform zero-shot transfer task for dialogue state tracking. We also employ TRADE \cite{wu-etal-2019-transferable} and SUMBT \cite{lee-etal-2019-sumbt} as baseline zero-shot learners for fair comparisons with the ATDM.

VHDA refers to model-based generation method using hierarchical variational autoencoder \cite{yoo-etal-2020-variational}. It generates dialogues incorporating information of speaker, goal of the speaker, turn-level dialogue acts, and utterance sequentially. \citet{yoo-etal-2020-variational} augment about 1,000 dialogues for restaurant and hotel domains in the MultiWOZ dataset. For a fair comparison, we use TRADE as the baseline model for the full data augmentation experiments. Also, we compare ours with the VHDA on the single-domain augmentation setting following their report.

\section{Experimental Results}

We use both joint goal accuracy (JGA) and slot accuracy (SA) as the performance measurement. The JGA is an accuracy which checks whether all slot values predicted at each turn exactly match the ground truth values, and the SA is the slot-wise accuracy of partial match against the grouth truth values. Especially for zero and few-shot setting, we follow the previous setup \cite{wu-etal-2019-transferable, campagna-etal-2020-zero}. Following \citet{campagna-etal-2020-zero}, the zero-shot learner model should be trained on data excluding the target domain, and tested on the target domain. We also add synthesized data from our NeuralWOZ which is trained in the same way, i.e., leave-one-out setup, to the training data in the experiment.

\subsection{Zero-Shot Domain Transfer Learning}

Our method achieves new state-of-the-art of zero-shot domain transfer learning for dialogue state tracking on the MultiWOZ 2.1 dataset (Table~\ref{table:zero-shot}). Except for the hotel domain, the performance over all target domains is significantly better than the previous sota method. We discuss the lower performance in hotel domain in the analysis section. Following the work of \citet{campagna-etal-2020-zero}, we also measure zero-shot coverage, which refers to the accuracy ratio between zero-shot learning over target domain, and fully trained model including the target domain. Our NeuralWOZ achieves 66.9\% and 79.2\% zero-shot coverage on TRADE and SUMBT, respectively, outperforming previous state-of-the-art, ATDM, which achieves 61.2\% and 73.5\%, respectively.

\subsection{Data Augmentation on Full Data Setting}

\begin{table}[t!]
    \centering
    \begin{threeparttable}
    \begin{tabular*}{0.85\columnwidth}{ccc}
        \toprule
        Synthetic & TRADE & SUMBT \\
        \midrule
        no syn & 44.2 / 96.5 & 46.7 / 96.7 \\
        ATDM & 43.0 / 96.4 & 46.9 / 96.6 \\
        NeuralWOZ & \textbf{45.8} / \textbf{96.7} & \textbf{47.1} / \textbf{96.8} \\ 
        \bottomrule
    \end{tabular*}
    \end{threeparttable}
    \caption{Full data augmentation on multi-domain DST. Joint goal accuracy / slot accuracy are reported.}
    \label{table:full-data-multi}
\end{table}

For full data augmentation, our synthesized data come from fully trained model including all five domains in this setting. Table \ref{table:full-data-multi} shows that our model still consistently outperforms in full data augmentation of multi-domain dialogue state tracking. Specifically, our NeuralWOZ performs 2.8\% point better on the joint goal accuracy of TRADE than ATDM. Our augmentation improves the performance by a 1.6\% point while ATDM degrades.

\begin{table}[t!]
    \centering
    \begin{threeparttable}
    \begin{tabular*}{0.85\columnwidth}{ccc}
        \toprule
        Synthetic & Restaurant & Hotel \\
        \midrule
        no syn & 64.1 / 93.1 & 52.3 / 91.9 \\
        VHDA & 64.9 / 93.4 & 52.7 / 92.0 \\
        NeuralWOZ & \textbf{65.8} / \textbf{93.6} & \textbf{53.5} / \textbf{92.1} \\ 
        \bottomrule
    \end{tabular*}
    \end{threeparttable}
    \caption{Full data augmentation on single-domain DST. Joint goal accuracy / slot accuracy are reported. TRADE is used for evaluation.}
    \label{table:full-data-single}
\end{table}

We also compare NeuralWOZ with VHDA, a previous model-based data augmentation method for dialogue state tracking \cite{yoo-etal-2020-variational}. Since the VHDA only considers single-domain simulation, we use single-domain dialogue in hotel and restaurant domains for the evaluation. Table \ref{table:full-data-single} shows that our method still performs better than the VHDA in this setting.
NeuralWOZ has more than twice better joint goal accuracy gain than that of VHDA.

\subsection{Intrinsic Evaluation of NeuralWOZ}

\begin{table}[t!]
    \centering
    \begin{threeparttable}
    \begin{tabular*}{0.85\columnwidth}{lcc}
        \toprule
        Domain & Collector $\downarrow$ & Labeler  $\uparrow$ \\
        \midrule
        
        Full & 5.0 & 86.8 \\
        \midrule
          w/o Hotel & 5.4 & 79.2 \\
          w/o Restaurant & 5.3 & 81.3 \\
          w/o Attraction & 5.3 & 83.4 \\
          w/o Train & 5.6 & 83.2 \\
          w/o Taxi & 5.2 & 83.1 \\
        \bottomrule
    \end{tabular*}
    \end{threeparttable}
    \caption{Intrinsic evaluation results of NeuralWOZ on the validation set of MultiWOZ 2.1. Perplexity and joint goal accuracy are used for measurement respectively. The ``w/o" means the domain is excluded from the full data. Different from the zero-shot experiments, the joint goal accuracy is computed by regarding all five domains.}
    \label{table:ablation-collectr-labeler}
\end{table}

Table \ref{table:ablation-collectr-labeler} shows the intrinsic evaluation results from two components (Collector and Labeler) of the NeuralWOZ on the validation set of MultiWOZ 2.1. We evaluate each component using perplexity for Collector and joint goal accuracy for Labeler, respectively. Note that the joint goal accuracy is achieved by using state candidate set, prepopulated as the multiple-choice options from the ground truth, $B_{1:T}$, as the training time of Labeler. It can be seen as using meta information since its purpose is accurate annotation but not the dialogue state tracking itself. We also report the results by excluding target domain from full dataset to simulate zero-shot environment. Surprisingly, synthesized data from ours performs effectively even though the annotation by Labeler is not perfect. We conduct further analysis, the responsibility of each model, in the following section.

\section{Analysis}
\subsection{Error Analysis}

\begin{figure}[t] 
\centering
\includegraphics[width=0.47\textwidth]{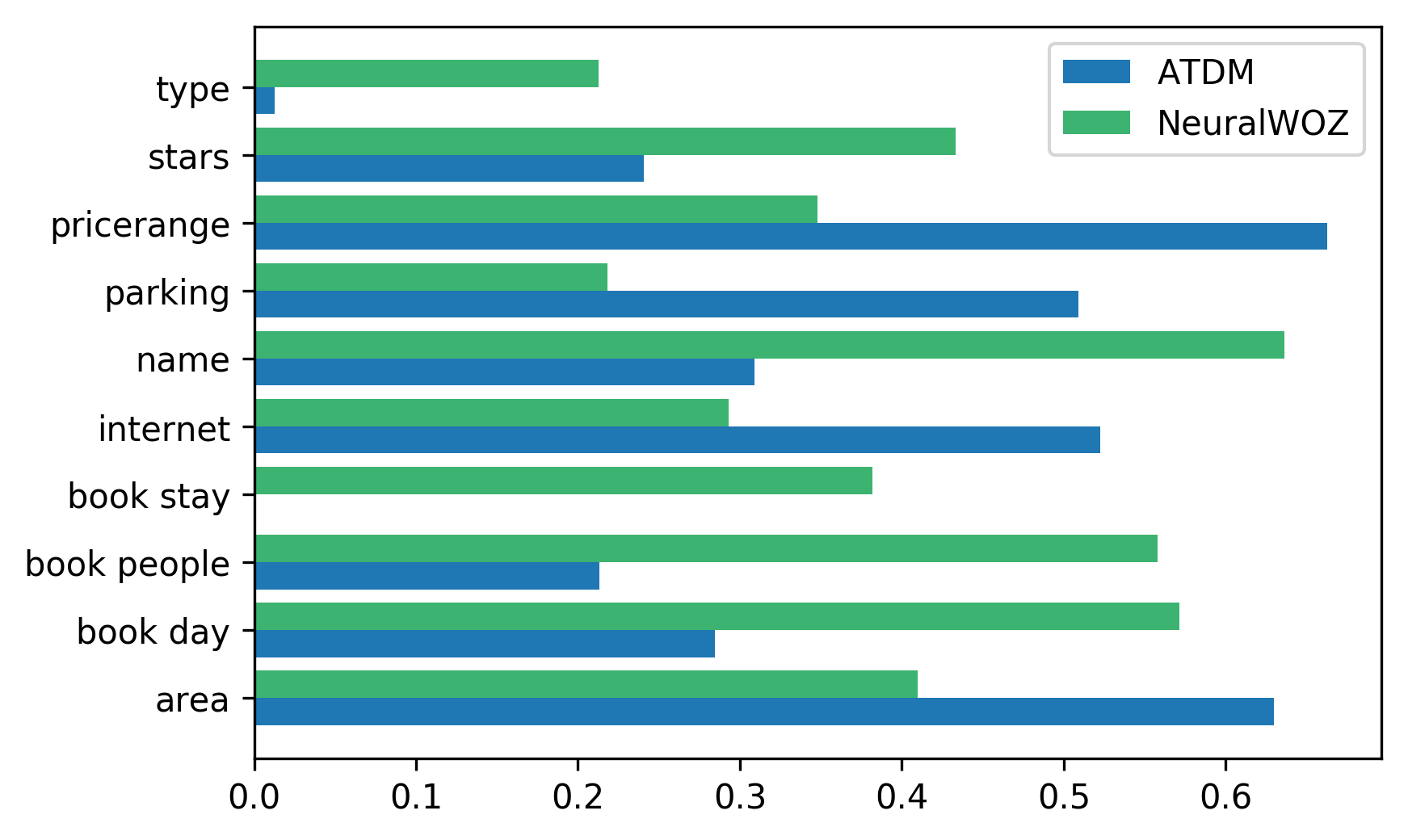}
\caption{Breakdown of accuracy by slot of hotel domain in the zero-shot experiments when using synthetic data. The analysis is conducted based on TRADE.}
\label{fig:error-zero-hotel}
\end{figure}

Figure \ref{fig:error-zero-hotel} shows the slot accuracy for each slot type in the hotel domain, which is the weakest domain from ours. Different from other four domains, only the hotel domain has two boolean type slots, ``parking" and ``internet", which can have only ``yes" or ``no" as their value. Since they have abstract property for the tracking, Labeler's labeling performance tends to be limited to this domain. However, it is noticeable that our accuracy of booking related slots (book stay, book people, book day) are much higher than the ATDM's. Moreover, the model using synthetic data from the ATDM totally fails to track the ``book stay" slot. In the synthesizing procedures of \citet{campagna-etal-2020-zero}, they create the data with a simple substitution of a domain noun phrase when the two domains have similar slots. For example, ``find me a restaurant in the city center'' can be replaced with ``find me a hotel in the city center'' since the restaurant and hotel domains share ``area'' slot.
We presume it is why they outperform over slots like ``pricerange" and ``area".

\subsection{Few-shot Learning}

\begin{figure}[t] 
\centering
\includegraphics[width=0.47\textwidth]{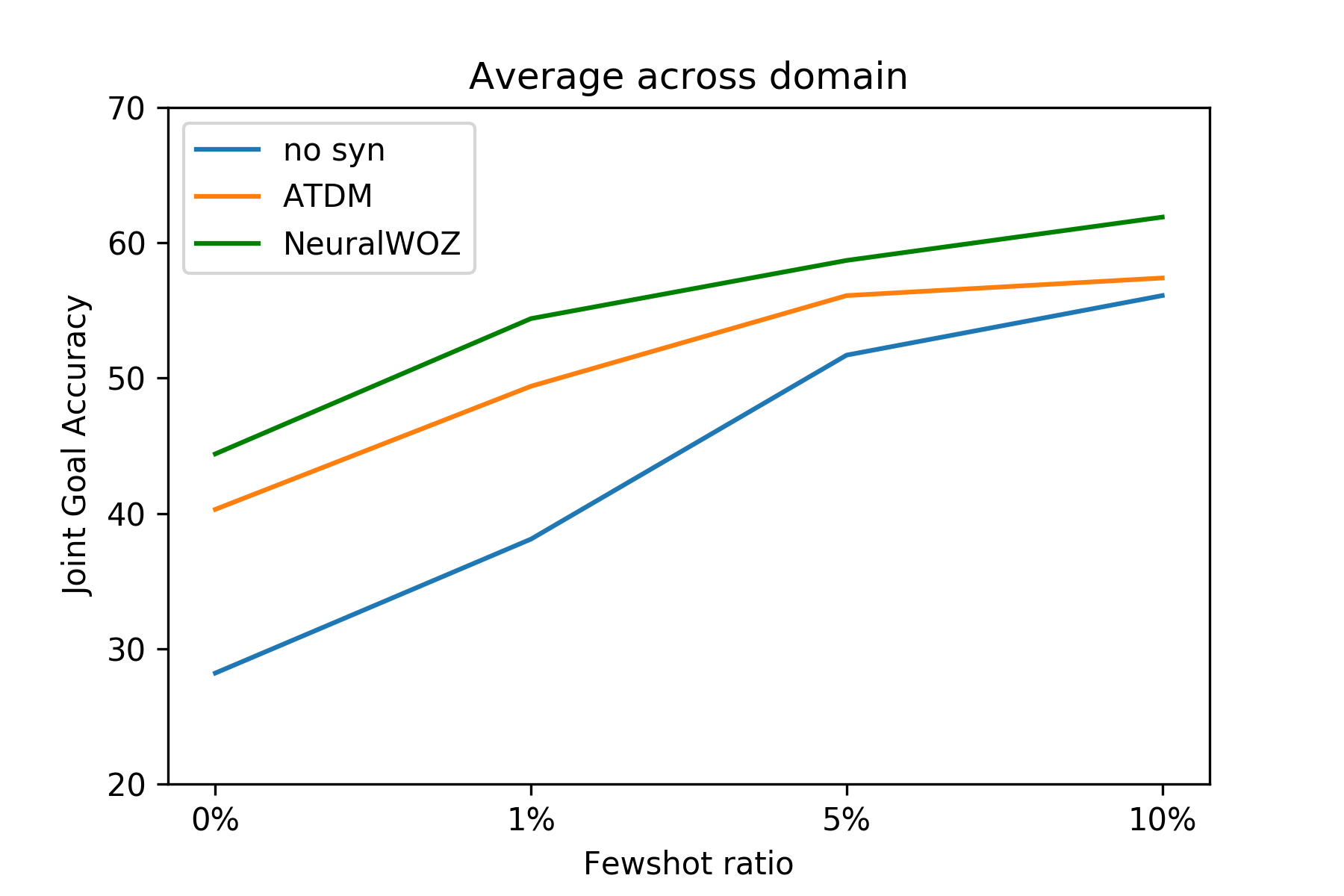}
\caption{Few-shot learning result in MultiWOZ 2.1. The score indicates average across domain. TRADE is used for the baseline model.}
\label{fig:few-shot-TRADE}
\end{figure}

We further investigate how our method is complementary with human-annotated data. Figure \ref{fig:few-shot-TRADE} illustrates our NeuralWOZ shows a consistent gain in the few-shot domain transfer setting. Unlike the performance with ATDM is saturated as few-shot ratio increases, the performance using our NeuralWOZ is improved continuously. We get about 5.8\% point improvement from the case which does not use synthetic data when using 10\% of human-annotated data for the target domain. It implies our method could be used more effectively with the human-annotated data in a real scenario.

\subsection{Ablation Study}

We discover whether Collector and Labeler are more responsible for the quality of synthesizing. Table \ref{table:responsibility} shows ablation results where each model of NeuralWOZ is trained the data including or withholding the hotel domain. Except for the training data for each model, the pipelined models are trained and dialogues are synthesized in the same way. Then, we train TRADE model using the synthesized data and evaluate it on hotel domain like the zero-shot setting. The performance gain from Collector which is trained including the target domain is 4.3\% point, whereas the gain from Labeler is only 0.8\% point. 
It implies the generation quality from Collector is more responsible for the performance of the zero-shot learner than accurate annotation of Labeler.

\begin{table}[t!]
    \centering
    \begin{threeparttable}
    \begin{tabular*}{0.84\columnwidth}{llc}
        \toprule
        Collector & Labeler & Hotel's JGA \\
        \midrule
        Full & Full & 53.5 \\
        Full & w/o Hotel & 30.8 \\
        w/o Hotel & Full & 27.3 \\
        w/o Hotel & w/o Hotel & 26.5 \\
        \bottomrule
    \end{tabular*}
    \end{threeparttable}
    \caption{Result of responsibility analysis. We compare the performances of each model with and without the hotel domain in the training data.}
    \label{table:responsibility}
\end{table}


\subsection{Qualitative Analysis}
\begin{figure}[t] 
\centering
\includegraphics[width=0.47\textwidth]{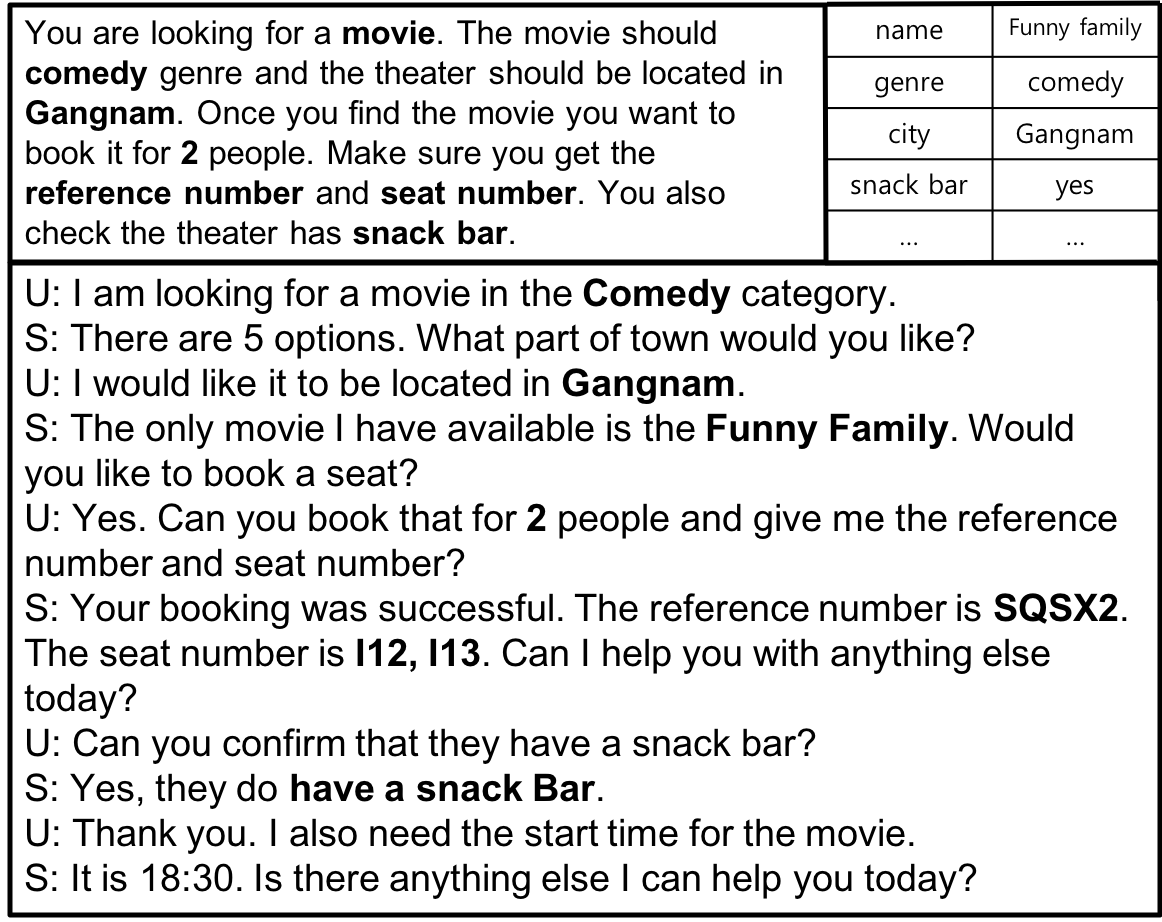}
\caption{Unseen domain dialogue generation from NeuralWOZ. The movie domain is an example. It has very different domain schema from the domains in MultiWOZ dataset.}
\label{fig:qualitative_unseen_1}
\end{figure}

Figure \ref{fig:qualitative_unseen_1} is an qualitative example generated by NeuralWOZ. It shows the NeuralWOZ can generate an unseen movie domain which has a different schema from the traveling, the meta domain of the MultiWOZ dataset, even if it is trained on only the dataset. It is harder to generalize when the schema structure of the target domain is different from the source domain. Other examples can be found in Appendix \ref{appendix.qualitative}. We would like to extend the NeuralWOZ to more challenging expansion scenario like these in future work.

\subsection{Comparison on End-to-End Task}
To show that our framework can be used for other dialogue tasks, we test our data augmentation method on end-to-end task in MultiWOZ 2.1. We describe the result in Appendix \ref{appendix.simulatedchat} with discussion. In full data setting, Our method achieves 17.46 BLUE, 75.1 Inform rate, 64.6 Success rate, and 87.31 Combine rate, showing performance gain using the synthetic data. 
Appendix \ref{appendix.simulatedchat} also includes the comparison and discussion on SimulatedChat \cite{mohapatra2020simulated}.

\section{Conclusion}
We propose NeuralWOZ, a novel dialogue collection framework, and we show our method achieves state-of-the-art performance on zero-shot domain transfer task. We find the dialogue corpus from NeuralWOZ is synergetic with human-annotated data. Finally, further analysis shows that NeuralWOZ can be applied for scaling dialogue system. We believe NeuralWOZ will spark further research into dialogue system environments where expansion target domains are distant from the source domains.

\section*{Acknowledgments}
We thank Sohee Yang, Gyuwan Kim, Jung-Woo Ha, and other members of NAVER AI for their valuable comments. We also thank participants who helped our preliminary experiments for building data collection protocol. 


\bibliographystyle{acl_natbib}
\balance
\bibliography{acl2021}

\clearpage
\appendix
\onecolumn

\section{Goal Instruction Sampling for Synthesizing in NeuralWOZ}
\label{appendix.synthesize}
\begin{figure}[h] 
\centering
\includegraphics[width=0.98\textwidth]{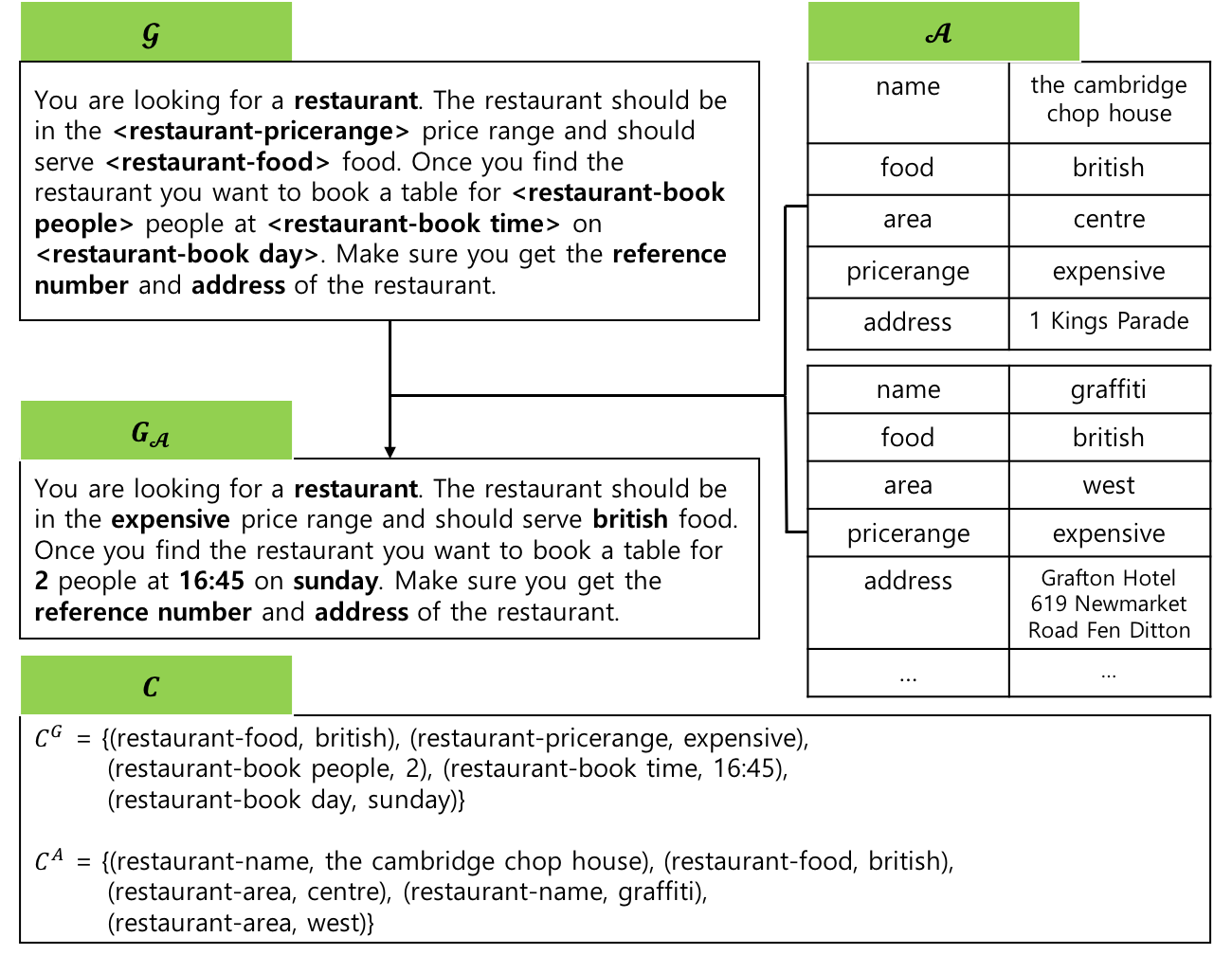}
\caption{An example of sampling goal instruction $G_{\mathcal{A}}$ using goal template $\mathcal{G}$ and randomly selected API call results $\mathcal{A}$.}
\label{fig:goal_sampling}
\end{figure}

\section{Data Statistics}
\label{appendix.stats}
\begin{table*}[!h]
    \centering
    \footnotesize
    \begin{threeparttable}
    \begin{tabularx}{0.98\textwidth}{llrrrrrr}
        \toprule
            & & \multicolumn{3}{c}{\# of Dialogues} & \multicolumn{3}{c}{\# of Turns} \\
        \cmidrule{3-8}
        Domain    & Slots & Train & Valid & Test & Train & Valid & Test \\
        \midrule
        Attraction & \makecell[Xt]{area, name, type} 
        & \quad 2,717 & 401 & 395 
        & \quad 8,073 & 1,220 & 1,256 \\
        \cmidrule{1-8}
        Hotel & \makecell[Xt]{price range, type, parking, book stay, book day, book people, area, stars, internet, name} 
        & \quad 3,381 & 416 & 394
        & \quad 14,793 & 1,781 & 1,756 \\
        \cmidrule{1-8}
        Restaurant & \makecell[Xt]{food, price range, area, name, book time, book day, book people} 
        & \quad 3,813 & 438 & 437
        & \quad 15,367 & 1,708 & 1,726 \\
        \cmidrule{1-8}
        Taxi & \makecell[Xt]{leave at, destination, departure, arrive by} 
        & \quad 1,654 & 207 & 195 
        & \quad 4,618 & 690 & 654 \\
        \cmidrule{1-8}
        Train & \makecell[Xt]{destination, day, departure, arrive by, book people, leave at}
        & \quad 3,103 & 484 & 494
        & \quad 12,133 & 1,972 & 1,976 \\
        \bottomrule
    \end{tabularx}
    \end{threeparttable}
    \caption{Data Statistics of MultiWOZ 2.1.}
    \label{table:data-statistics}
\end{table*}

\begin{table*}[t!]
    \centering
    \begin{threeparttable}
    \begin{tabular*}{0.98\textwidth}{lrrrrrr}
        \toprule
        & \textbf{Attraction} & \textbf{Hotel} & \textbf{Restaurant} & \textbf{Taxi} & \textbf{Train}  & \textbf{Full} \\
        \hline
        \# goal template & 411 & 428 & 455 & 215 & 482 & 1,000 \\
        \# synthesized dialogues & 5,000 & 5,000 & 5,000 & 5,000 & 5,000 & 1,000 \\
        \# synthesized turns & 38,655 & 38,112 & 37,230 & 45,542 & 37,863 & 35,053 \\
        \# synthesized tokens & 947,791 & 950,272 & 918,065 & 1,098,917 & 873,671 & 856,581 \\
        \bottomrule
    \end{tabular*}
    \end{threeparttable}
    \caption{Statistics of the synthesized data used in NeuralWOZ using for zero-shot and full augmentation experiments.}
    \label{table:}
\end{table*}

\section{Additional Qualitative Examples}
\label{appendix.qualitative}
\begin{figure}[ht!] 
\centering
\includegraphics[width=0.98\textwidth]{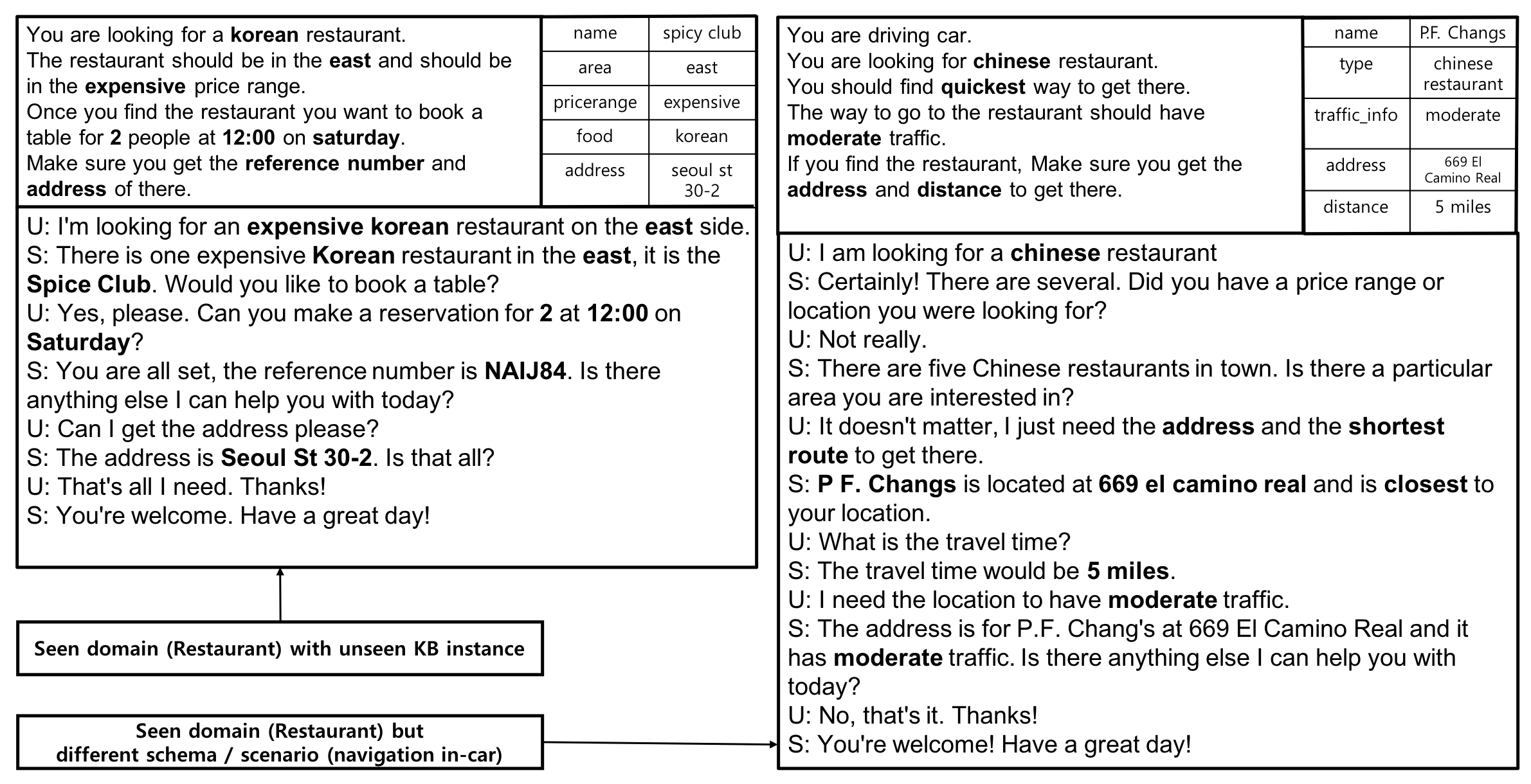}
\caption{Qualitative examples of synthesized dialogues from NeuralWOZ in the restaurant domain.}
\label{fig:qualitative_seen_1}
\end{figure}

 Figure \ref{fig:qualitative_seen_1} shows other examples from our NeuralWOZ. The left subfigure shows an example of synthesized dialogue from NeuralWOZ in a restaurant, which is seen domain and has the same schema from the restaurant domain in MultiWOZ dataset. However, the ``spicy club'' is an unseen instance which is newly added to the schema for the synthesizing. The right subfigure shows other synthetic dialogue in restaurant, which is a seen domain but has different schema from restaurant domain in MultiWOZ dataset. It describes navigation in-car scenario which is borrowed from KVret dataset \cite{eric2017keyvalue}. It is a non-trivial problem to adapt to unseen scenario, even if it is in the same domain.

\section{Additional Explanation on Comparison in End-to-End Task}
\label{appendix.simulatedchat}

\begin{table*}[t!]
    \centering
    \footnotesize
    \begin{threeparttable}
    \begin{tabular*}{0.87\textwidth}{ccccccc}
        \toprule
        Model & Belief State & BLEU & Inform & Success & Combined \\
        \midrule
        DAMD \cite{zhang2020task} & Oracle & 17.3 & 80.3 & 65.1 & 90\\
        SimpleTOD \cite{hosseiniasl2020simple} & Oracle & 16.22 & 85.1 & 73.5 & 95.52\\
        GPT2 \cite{mohapatra2020simulated} & Oracle & 15.95 & 72.8 & 63.7 & 84.2\\
        GPT2 + SimulatedChat \cite{mohapatra2020simulated} & Oracle & 15.06 & 80.4 & 62.2 & 86.36\\
        GPT2 (ours) & Oracle & 17.27 & 77.1 & 67.8 & 89.72\\
        GPT2 + NeuralWOZ (ours) & Oracle & 17.69 & 78.1 & 67.6 & 90.54\\ 
        \midrule
        DAMD \cite{zhang2020task} & Generated & 18.0 & 72.4 & 57.7 & 83.05\\
        SimpleTOD \cite{hosseiniasl2020simple} & Generated & 14.99 & 83.4 & 67.1 & 90.24\\
        GPT2 \cite{mohapatra2020simulated} & Generated & 15.94 & 66.2 & 55.4 & 76.74\\
        GPT2 + SimulatedChat \cite{mohapatra2020simulated} & Generated & 14.62 & 72.5 & 53.7 & 77.72\\ 
        GPT2 (ours) & Generated & 17.38 & 74.6 & 64.4 & 86.88\\
        GPT2 + NeuralWOZ (ours) & Generated & 17.46 & 75.1 & 64.6 & 87.31\\ 
        
        \bottomrule
    \end{tabular*}
    \end{threeparttable}
    \caption{Performance of the end-to-end task model.}
    \label{table:end-to-end}
\end{table*}

To compare our model with the model of \cite{mohapatra2020simulated}, we conduct end-to-end task experiments the previous work did. Table \ref{table:end-to-end} illustrates the result.
Though the performance of baseline implementation is different, we can see that the trend of performance improvement is comparable to the report of SimulatedChat.


Two studies are also different in terms of modeling. In our method, all utterances in the dialogue are first collected based on goal instruction and KB information by Collector. After that, Labeler selects annotations from candidate labels, which can be inducted from goal instruction and KB information. On the other hand, SimulatedChat creates utterance and label sequentially with knowledge base access, for each turn. Thus, each generation of utterance is affected by the generated utterance of labels of the previous turn.

In detail, the two methods also differ in terms of complexity. SimulatedChat creates a model for each domain separately, and for each domain, it creates five neural modules: user response generation, user response selector, agent query generator, agent response generator, and agent response selector.
This results 25 neural models for data augmentation in the MultiWOZ experiments. On the contrary, NeuralWOZ only needs two neural models for data augmentation: Collector and Labeler.

Another notable difference is that SimulatedChat does not generate multi-domain data in a natural way. The strategy of creating a model for each domain not only makes it difficult to transfer the knowledge to a new domain, but also makes it difficult to create multi-domain data.
In SimulatedChat, the dialogue is created for each domain and then concatenated.
Our model can properly reflect the information of all domains included in the goal instruction to generate synthetic dialogues, regardless of the number of domains.

\section{Other Experiment Details}
\label{appendix.training_details}

The number of parameters of our models is 406M for Collector and 124M for Labeler, respectively. Both models are trained on two V100 GPUs with mixed precision floating point arithmetic. It takes about 4 (10 epochs) and 24 hours (30 epochs) for the training, respectively. We optimize hyperparameters of each model, learning rate \{1e-5, 2e-5, 3e-5\} and batch size \{16, 32, 64\}, based on greedy search. We set the maximum sequence length of Collector to 768 and the Labeler to 512.

For the main experiments, we fix hyperparameter settings of TRADE (learning rate 1e-4 and batch size 32) and SUMBT (learning rate 5e-5 and batch size 4) same with previous works. We use the script of \citet{campagna-etal-2020-zero} for converting the TRADE's data format to the SUMBT's.

For GPT2 \cite{radford2019language} based model for the end2end task, we re-implement the model similar with SimpleTOD \cite{hosseiniasl2020simple} but not using action. Thus, it generates dialogue context, dialogue state, database results, and system response in an autoregressive manner. We also use special tokens in the SimpleTOD (without special tokens for the action). We follow preprocessing procedure for the end2end task, including delexicalization suggested by \cite{budzianowski-etal-2018-multiwoz}. We use 8 for batch size and 5e-5 for learning rate. Note that we also train our NeuralWOZ using 30\% of training data and synthesize 5000 dialogues for the end2end experiments. However, we could not find detailed experiments setup of \citet{mohapatra2020simulated} including hyperparameter, the seed of each portion of training data, and evaluation, so it is not a fair comparison.

\end{document}